\documentclass[10pt,twocolumn,letterpaper]{article}

\usepackage{amsmath,amsfonts,bm}

\def\eqref#1{equation~\ref{#1}}

\def\1{\bm{1}}

\def\rve{{\mathbf{e}}}

\def\rvu{{\mathbf{i}}}

\def\rvp{{\mathbf{p}}}

\def\rvu{{\mathbf{u}}}
\def\rvv{{\mathbf{v}}}

\def\rvx{{\mathbf{x}}}

\DeclareMathAlphabet{\mathsfit}{\encodingdefault}{\sfdefault}{m}{sl}
\SetMathAlphabet{\mathsfit}{bold}{\encodingdefault}{\sfdefault}{bx}{n}

\def\gD{{\mathcal{D}}}

\def\gF{{\mathcal{F}}}

\def\gP{{\mathcal{P}}}

\def\gR{{\mathcal{R}}}
\def\gS{{\mathcal{S}}}

\def\gX{{\mathcal{X}}}
\def\gY{{\mathcal{Y}}}

\def\sR{{\mathbb{R}}}

\newcommand{\R}{\mathbb{R}}

\usepackage{iccv}              
\usepackage{amsmath}
\usepackage{amssymb}
\usepackage{amsfonts}
\usepackage{multirow}
\usepackage{bm}
\usepackage{dsfont}
\usepackage{graphicx}
\usepackage{cases}

\usepackage{subcaption}  
\usepackage{caption}
\usepackage[misc]{ifsym}
\usepackage[T1]{fontenc}
\usepackage[accsupp]{axessibility}  
\definecolor{iccvblue}{rgb}{0.21,0.49,0.74}
\usepackage[pagebackref,breaklinks,colorlinks,allcolors=iccvblue]{hyperref}

\def\paperID{4413} 
\def\confName{ICCV}
\def\confYear{2025}

\title{Joint Asymmetric Loss for Learning with Noisy Labels}

\author{Jialiang Wang\quad Xianming Liu\thanks{Corresponding author} \quad Xiong Zhou \quad Gangfeng Hu\quad Deming Zhai \quad Junjun Jiang \\
Harbin Institute of Technology \vspace{4pt}\\
Xiangyang Ji \\
Tsinghua University \\
}

\newtheorem{Theorem}{Theorem}[section] 
\newtheorem{Definition}{Definition}[section]

\begin{document}
\maketitle
\begin{abstract}
Learning with noisy labels is a crucial task for training accurate deep neural networks.
To mitigate label noise, prior studies have proposed various robust loss functions, particularly symmetric losses. Nevertheless, symmetric losses usually suffer from the underfitting issue due to the overly strict constraint. To address this problem, the Active Passive Loss (APL) jointly optimizes an active and a passive loss to mutually enhance the overall fitting ability.
Within APL, symmetric losses have been successfully extended, yielding advanced robust loss functions.
Despite these advancements, emerging theoretical analyses indicate that asymmetric losses,  a new class of robust loss functions, possess superior properties compared to symmetric losses. 
However, existing asymmetric losses are not compatible with advanced optimization frameworks such as APL, limiting their potential and applicability. Motivated by this theoretical gap and the prospect of asymmetric losses, 
we extend the asymmetric loss to the more complex passive loss scenario and propose the Asymetric Mean Square Error (AMSE), a novel asymmetric loss. We rigorously establish the necessary and sufficient condition under which AMSE satisfies the asymmetric condition.
By substituting the traditional symmetric passive loss in APL with our proposed AMSE, we introduce a novel robust loss framework termed Joint Asymmetric Loss (JAL).
Extensive experiments demonstrate the effectiveness of our method in mitigating label noise.  Code available at: \url{https://github.com/cswjl/joint-asymmetric-loss}

\end{abstract}    
\section{Introduction}
\label{sec:intro}
\begin{figure}[!t]
    \centering

    \begin{subfigure}{0.235\textwidth}  
        \centering
        \includegraphics[width=\textwidth]{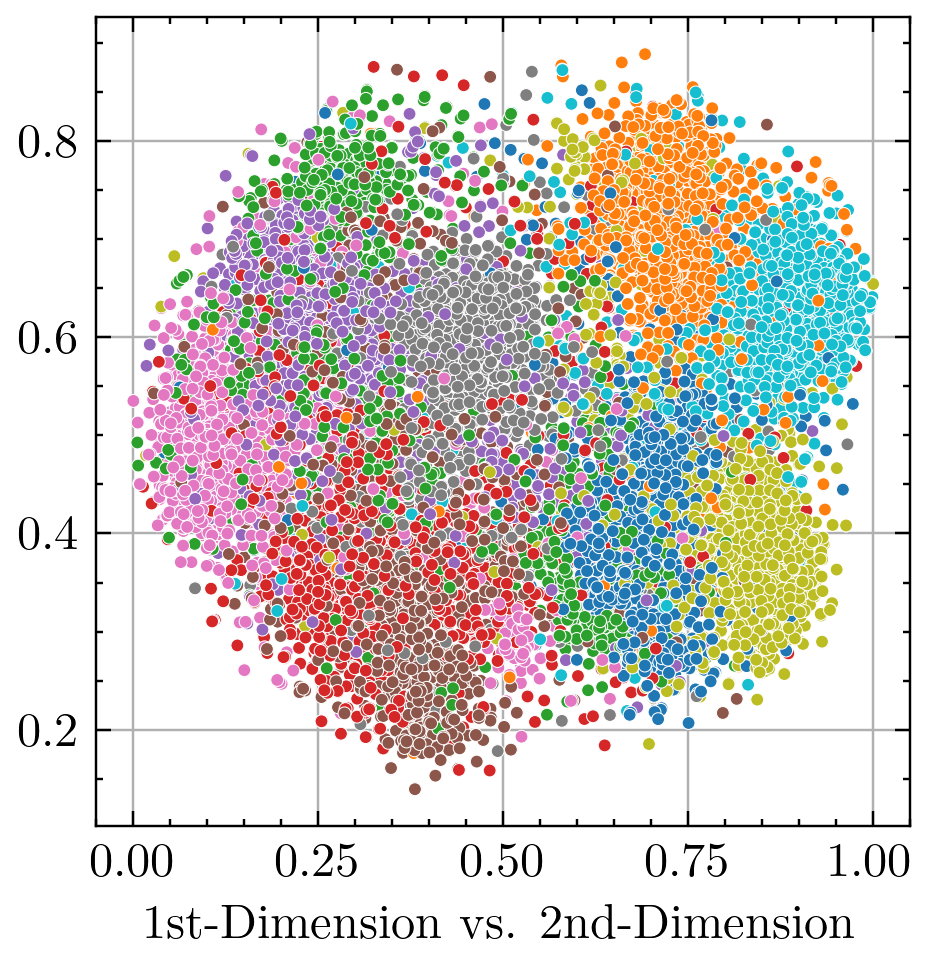}
        \caption{CE}
        \label{fig: tsne_ce}
    \end{subfigure}
    \hfill
    \begin{subfigure}{0.235\textwidth}
        \centering
        \includegraphics[width=\textwidth]{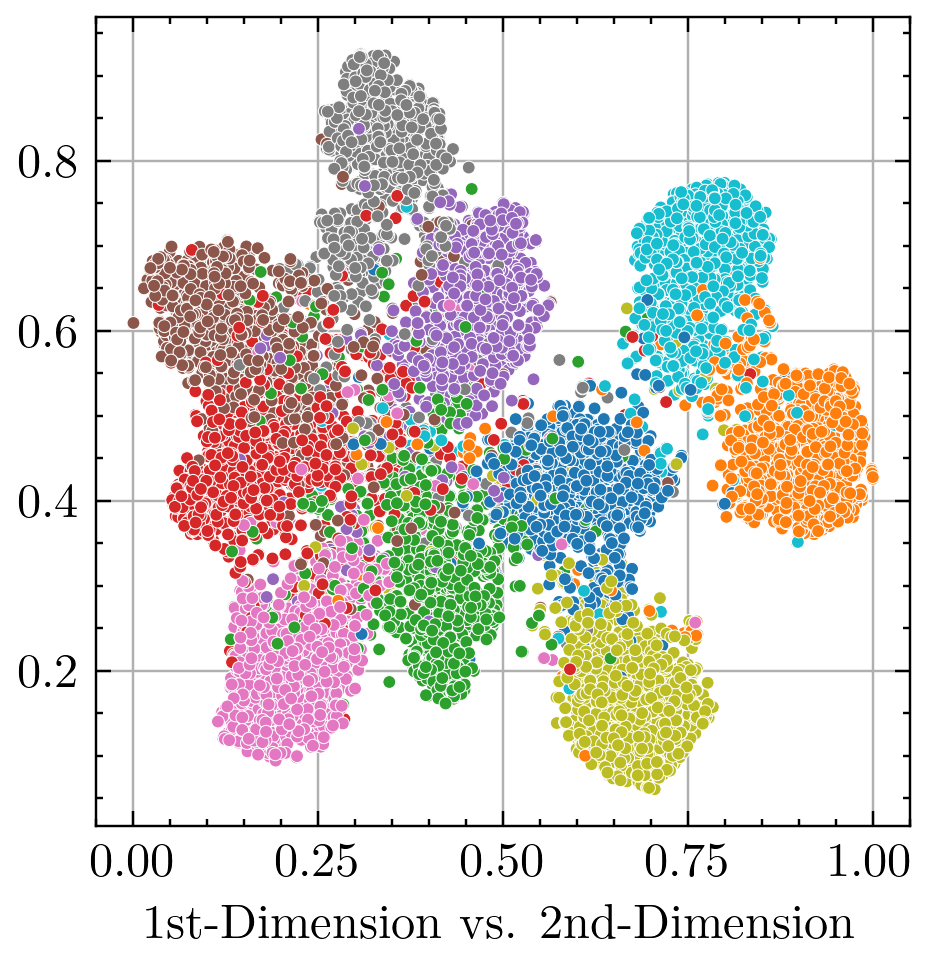}
        \caption{JAL-CE}
        \label{fig: tsne_jal_ce}
    \end{subfigure}
    \\
    \begin{subfigure}{0.235\textwidth}
        \centering
        \includegraphics[width=\textwidth]{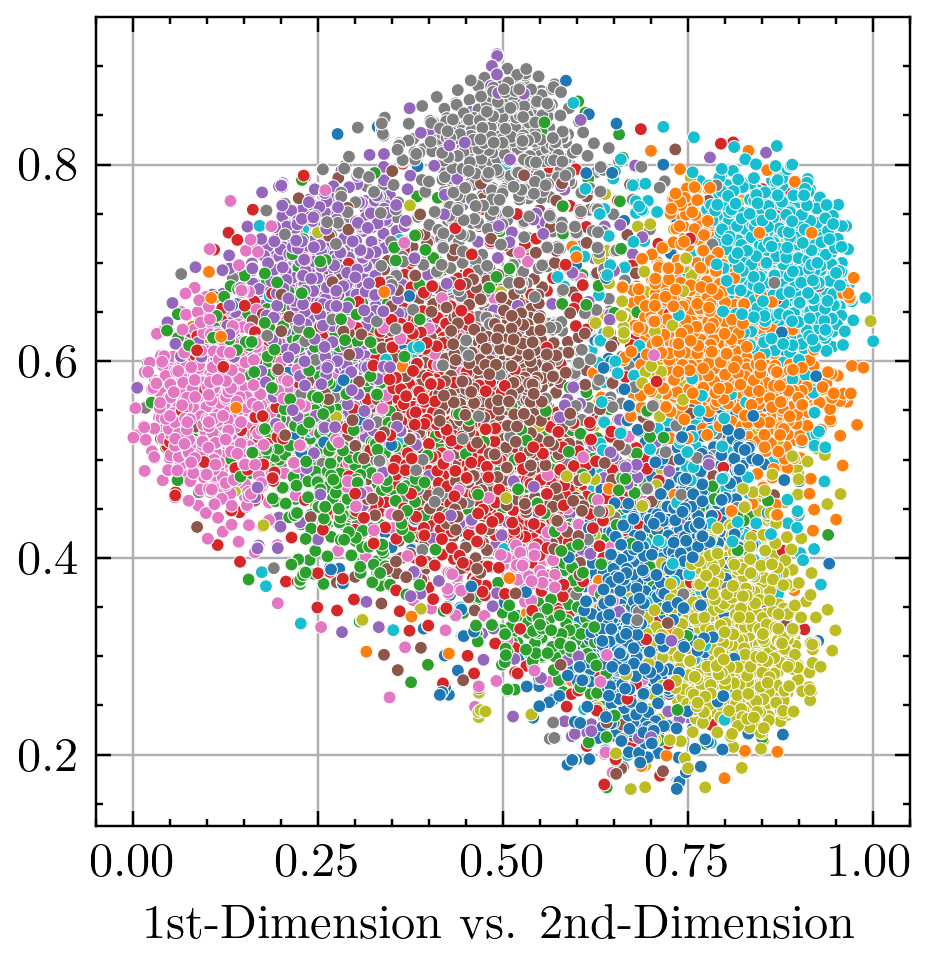}
        \caption{FL}
        \label{fig: tsne_fl}
    \end{subfigure}
    \hfill
    \begin{subfigure}{0.235\textwidth}
        \centering
        \includegraphics[width=\textwidth]{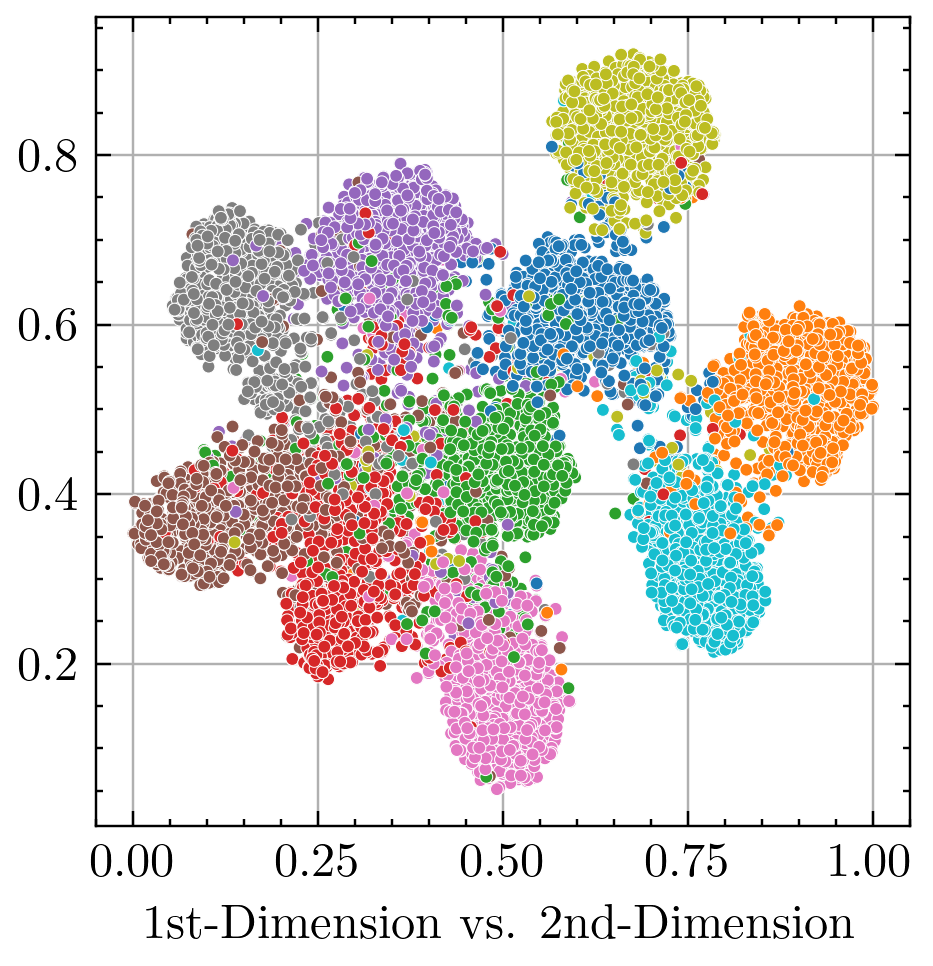}
        \caption{JAL-FL}
        \label{fig: tsne_jal_fl}
    \end{subfigure}

    \vskip-5pt
    \caption{Visualizations of 2D t-SNE \citep{tsne} embeddings of learned representations on the CIFAR-10 test set, from models trained with 0.4 symmetric noise.
    The representations learned by the proposed JAL method are with more separated and clearly bound margin.
    }
    \label{fig: tsne}
\end{figure}

Deep neural networks (DNNs) have demonstrated outstanding performance in a wide range of machine learning tasks \cite{deep_learning, lnl_survey}. However, the prevalence of noisy labels in real-world datasets remains a significant challenge, often arising from human carelessness or a lack of domain expertise \cite{lnl_survey}.
Applying supervised learning methods directly to noisy labeled data typically degrades model performance \cite{arpit2017closer}. Furthermore, the ability to generalize from noisy supervision is crucial for aligning large language models \cite{weak2strong}. As a result, developing noise-tolerant learning techniques has become a critical and increasingly studied problem within weakly supervised learning.
Among various approaches proposed in the literature, designing robust loss functions has gained particular popularity due to its simplicity and broad applicability \cite{Ghosh2017robust, NCE, ALFs_ICML, student_loss}.

Previous works \citep{manwani2013noise, van2015learning, Ghosh2017robust} theoretically proved that symmetric loss functions are inherently tolerant to label noise under some moderate assumptions. However, the fitting ability of symmetric loss functions is constrained by the overly strict symmetric condition \citep{ALFs_ICML}. 
Symmetric loss functions such as Mean Absolute Error (MAE) \citep{Ghosh2017robust} have proven challenging to optimize. 
To address this underfitting issue, inspired by complementary learning \cite{NLNL}, \citet{NCE} proposed the Active Passive Loss (APL) framework. 
They categorize loss functions into two types: 1)``Active loss", which only explicitly maximizes the probability of the labeled class, and 2) ``Passive loss", which also explicitly minimizes the probabilities of other classes.
APL simultaneously employs an active loss and a passive loss to enhance each other's optimization processes, improving overall fitting performance. By incorporating symmetric losses within the APL framework, several advanced robust loss functions have been developed \cite{NCE, ANL}.


Recently, \citet{ALFs_ICML, ALF_PAMI} proposed a novel class of robust loss functions called Asymmetric Loss Functions (ALFs). Their theoretical analysis shows that ALFs offer noise-tolerance to label noise under a more relaxed condition compared to symmetric loss functions.
However, existing asymmetric loss functions, such as Asymmetric Unhinged Loss (AUL) , are all active losses, as achieving the asymmetric condition for passive losses remains a challenging problem. 
Unfortunately, our explorations indicate that these existing asymmetric loss functions are not compatible with the APL framework.
The absence of a theoretical foundation for asymmetric loss functions in the passive loss scenario makes them unsuitable for the APL framework, thereby limiting their potential and practical applications.


In this paper, we extend asymmetric losses to the passive loss scenario, which is more challenging to analyze. We propose a new asymmetric passive loss function, called \textit{Asymmetric Mean Square Error} (AMSE). Our proposed AMSE is both simple and theoretically sound, and we rigorously establish the necessary and sufficient condition for it to satisfy the asymmetric condition. 
By replacing the traditional symmetric loss in APL with our proposed AMSE, we introduce a new framework called \textit{Joint Asymmetric Loss} (JAL). Our JAL enhances the traditional APL framework while preserving the complete noise-tolerance. 
Our key contributions are highlighted as follows:
\\
\begin{itemize}
\item We extend asymmetric losses to the more challenging passive loss scenario and propose a novel asymmetric loss function, \textit{Asymmetric Mean Square Error} (AMSE). Additionally, we rigorously establish the necessary and sufficient conditions for AMSE to satisfy the asymmetry condition.

\item 
By incorporating the proposed AMSE into the APL framework, we introduce a novel approach called \textit{Joint Asymmetric Loss} (JAL), which ensures robustness and enhances sufficient learning.

\item We conducted comprehensive ablation and comparison experiments. The extensive results highlight the superiority of our method.

\end{itemize}
\section{Related Work}
\label{sec: related work}

Learning with noisy labels, or called noise-tolerant learning, aims to train a robust model in the presence of noisy labels. Our paper concentrates on one prevalent research avenue: designing robust loss functions.

\citet{manwani2013noise, van2015learning, Ghosh2017robust} theoretically demonstrated that a loss function would be inherently tolerant to label noise as long as it satisfies the symmetric condition.
However, symmetric loss functions are difficult to optimize due to the over-strictmetric condition, such as Mean Absolute Error (MAE). 
This drawback motivates some works to combine the robust MAE with the well-fitting Cross Entropy (CE). Examples of such mixture loss functions include Generalized Cross Entropy (GCE) \citep{GCE}, Symmetric Cross Entropy (SCE) \citep{SCE}, Taylor Cross Entropy (Taylor-CE) \citep{tayor-ce}, and Jensen-Shannon Divergence Loss (JS) \citep{JS}.
These mixture loss functions often select an intermediate value between the gradients of CE and MAE, representing a trade-off between fitting ability and robustness. 
Sparse Regularization (SR) \cite{SR} and $\epsilon$-Softmax \cite{wangepsilon} approximate one-hot vectors to achieve a relaxed symmetric condition. 
Inspired by the complementary learning (NLNL \cite{NLNL} and JNPL \cite{JPNL}), Active Passive Loss (APL) \citep{NCE} and Active Negative Loss (ANL) \citep{ANL}, use two different symmetric losses simultaneously to improve the fitting ability. 
Recently, \citet{ALFs_ICML,ALF_PAMI} proposed a new family of robust loss functions for clean-label-dominant noise, namely asymmetric loss functions (ALFs). ALFs demonstrated better performance compared to symmetric loss functions. \citet{wei2022smooth} proposed a new label smoothing method, called Negative Label Smoothing (Negative-LS), improving robustness when learning with noisy labels.
In addition, PHuber-CE \cite{PHuber-CE} and LogitClip (LC) \cite{LC} mitigate the memorization of noisy labels by clamping the gradient and logit, respectively.

\section{Preliminary}
\noindent\textbf{Problem Definition.}
Considering a classification problem, we denote $\gX\subset \R^d$ as the sample space and $\gY=[K]=\{1,2,...,K\}$ as the label space, where $K$ is the number of classes. 
In the supervised scenario, a labeled dataset $\gS=\{(\rvx_n,y_n)\}_{n=1}^N$ is typically available for training classifiers, where  $(\rvx_n,y_n)$ are i.i.d draws from an underlying distribution $\gD$ over $\gX\times \gY$. 
The classifier $f: \gX \to \gP$ is a model with a softmax layer that maps the sample space $\gX$ to the probability simplex $\gP$, where $\gP = \left\{ \rvp \in [0,1]^K \mid \bm{1}^\top \rvp = 1 \right\}.$
The predicted label is then given by  $\hat{y} = \arg\max_k f(\rvx)_k.$
Moreover, let $L:\gP \times \gY \rightarrow \sR$ represent the classification loss function $L(f(\rvx),\rve_y)$, where $\rve_y$ is the one-hot vector with its $y$-th element set to 1. 
In this paper, we consider the loss functional, $L(\rvu,\rvv)=\sum_{k=1}^K\ell(u_k,v_k)$ with a basic loss function $\ell$, where $u_k$ is the k-th element of the vector
$\rvu$. 
For the sake of brevity, we abbreviate $L(f(\rvx),\rve_k)$ as $L(f(\rvx),k)$  in the following.

\vspace{4pt}
\noindent\textbf{Label Noise Model.} 
In the context of learning with noisy labels, we have access to a noisy training set 
$
\tilde{\mathcal{S}} = \{(\mathbf{x}_n, \tilde{y}_n)\}_{n=1}^N$
instead of its clean counterpart, \(\mathcal{S}\). For a given sample \(\rvx\), the noise corruption process is characterized by the flipping of the true label \(y\) into the observed label \(\tilde{y}\) with a conditional probability as follows:
\begin{equation}
\label{eq:symmetric noise}
 \tilde{y}=
\begin{cases}
    y &\text{with probability} \quad \eta_{\rvx,y}=1-\eta_\rvx\\ 
    k, k \in [K], k\neq y  &\text{with probability}\quad \eta_{\rvx,k}
\end{cases},
\end{equation}
where the overall noise rate for \(\rvx\) is given by 
$\eta_\rvx = \sum_{k \neq y} \eta_{\rvx, k}.$

Following previous works~\cite{Ghosh2017robust, IDN-PDN, NCE, ANL}, we primarily focus on three prevalent types of label noise: 
1) Symmetric Noise:
$\eta_{\rvx,y}=1-\eta$ and  $\eta_{\rvx,k \neq y}=\frac{\eta}{K-1}$, where noise rate $\eta_\rvx = \eta$ is a constant for any instance.
2) Asymmetric Noise:
$\eta_{\rvx,y}=1-\eta_y$ and    $\sum_{k \neq y}\eta_{\rvx,k}=\eta_y$, where $\eta_\rvx = \eta_y$  denotes the noise rate for the instance of $y$-th class.
3) Instance-Dependent Noise:
$\eta_{\rvx,y}=1-\eta_\rvx$ and $\sum_{k \neq y}\eta_{\rvx,k}=\eta_\rvx$, where $\eta_\rvx$ denotes the noise rate for the instance $\rvx$.
Herein, for asymmetric and instance-dependent noise, $\eta_{\rvx, i}$ is not necessarily equal to $\eta_{\rvx, j}$ for $i\neq j$.

\vspace{4pt}
\noindent\textbf{Risk Minimization and Noise-Tolerant Learning.}
In the case of clean labels, the expected risk~\cite{bartlett2006risk_bound} for a given loss function \(L\) and prediction function \(f\) is defined as  
$
\gR_L(f) = \mathbb{E}_{(\rvx,y) \sim \gD} [L(f(\rvx), y)].
$  
The goal of supervised learning is to find the expectation risk minimizer:  
$
f^* \in \arg\min_{f \in \gF} \gR_L(f).
$  
However, in the presence of noisy labels, we instead minimize the noisy risk, given by  
\begin{equation}
\label{noisy-L-risk} 
\gR_L^\eta(f) = \mathbb{E}_{\gD} [ (1 - \eta_{\rvx}) L(f(\rvx), y) + \sum_{k \neq y} \eta_{\rvx,k} L(f(\rvx), k) ],
\end{equation}  
where the term \(\sum_{k \neq y} \eta_{\rvx,k} L(f(\rvx), k)\) represents the noisy component, which often poses challenges in training deep neural networks (DNNs).  
As discussed in~\cite{Ghosh2017robust}, a loss function \(L\) is said to be \textit{noise-tolerant} if the global minimizer of the noisy risk,  
$
f^*_{\eta} \in \arg\min_f \gR_L^\eta(f),
$  
also minimizes the clean risk, i.e.,  
$
f^*_{\eta} \in \arg\min_f \gR_L(f).
$
\section{Methodology}
In this section, we first introduce the Active Passive Loss (APL) \cite{NCE} and Asymmetric Loss Functions (ALFs) \cite{ALFs_ICML, ALF_PAMI}, which are relevant to our work. We then present the proposed Asymmetric Mean Square Error (AMSE) and Joint Asymmetric Loss (JAL), followed by a rigorous theoretical analysis. 
\subsection{Active Passive Loss}

Previous works \citep{van2015learning, Ghosh2017robust} theoretically proved that a loss function is noise-tolerant to symmetric and asymmetric label noise under some mild assumptions if it is symmetric.
\begin{Definition}[Symmetric Condition]
\label{defin: symmetric condition}
A loss function $L$ is symmetric if it satisfies
\begin{equation}
    \sum_{k=1}^K L(f(\rvx), k) = C,
\end{equation}
where $C$ is a constant and $k \in [K]$ is the label corresponding to each class.
\end{Definition}

Based on this, \citet{NCE} proposed the normalized loss functions, which normalize a loss function by:
\begin{equation}
    L_\text{norm} = \frac{L(f(\rvx), y)}{\sum_{k=1}^KL(f(\rvx), k)}.
\end{equation}

This simple normalization operation can make any loss function symmetric, since we always have $ \sum_{k=1}^KL_\text{norm}(f(\rvx), k) = 1$. By normalizing Cross Entropy (CE) and Focal Loss (FL) \cite{FL}, \citet{NCE} proposed Normalized Cross Entropy (NCE) and Normalized Focal Loss (NFL). However, similar to symmetirc MAE, both NCE and NFL are challenging to optimize due to the overly strict symmetric condition. To address this issue, \citet{NCE} characterize existing loss functions into two types: \textit{Active} and \textit{Passive}. For a loss $L(f(\rvx), y) = \sum_{k=1}^K\ell(f(\rvx)_k,e_ k)$, where $f(\rvx)_k$ is the k-th element of the  prediction vector
$f(\rvx) = \rvp(\cdot|\rvx)$ and $e_k$ is the k-th element of the label 
$\rve_y$ (e.g., for CE loss, we have $L (f(\rvx), y) = \sum_{k=1}^K-e_k\log f(\rvx)_k$), we have the following definitions \cite{NCE, ANL}:
\begin{Definition}[Active Loss Function]
\label{defin: active loss}
$L_\text{active}$ is an active loss function if  $\forall(\rvx, y)\in \gD, \forall k \neq y, \ell(f(\rvx)_k, e_k) = 0$. 
\end{Definition}

\begin{Definition}[Passive Loss Function]
\label{defin: passive loss}
$L_\text{passive}$ is a passive loss function if  $\forall(\rvx, y)\in \gD, \exists k \neq y, \ell(f(\rvx), e_k) \neq 0$. 
\end{Definition}

According to definitions, active loss functions only explicitly maximize classifier's output probability at the class position specified by the label $y$. In contrast, passive loss functions also explicitly minimize the probability at least one other class positions.
The active loss functions include CE, FL, NCE/NFL \cite{NCE}, while the passive loss functions include MAE, and NNCE/NNFL \cite{ANL}\footnote{The active and passive definitions and the type of loss functions reference \cite{NCE, ANL}.  }.

To address the underfitting issue of symmetric losses, \citet{NCE} proposed the Active Passive Loss (APL):
\begin{equation}
    L_\text{APL} = \alpha \cdot L_\text{active} + \beta \cdot L_\text{passive},
\end{equation}
where $\alpha, \beta >0$ are parameters.
By combining the two different symmetric loss functions, APL can improve the fitting ability under the premise of ensuring robustness.
Through combining active NCE/NFL and passive MAE, \citet{NCE} get one of the state-of-the-art methods. 

Additionally, \citet{ANL} proposed new passive symmetric loss functions, known as Normalized Negative Loss Functions (NNCE/NNFL).
By replacing the MAE in APL with  NNCE/NNFL, they proposed a new method, named  Active Negative Loss (ANL).  
However, both APL \cite{NCE} and ANL \cite{ANL} are limited to symmetric loss functions within the APL framework. To date, no research has explored the potential benefits of incorporating higher-performing asymmetric loss functions \cite{ALFs_ICML, ALF_PAMI} into the APL framework.

\subsection{Asymmetric Loss Functions}
Recently, \citet{ALFs_ICML, ALF_PAMI} proposed a new class of robust loss functions, called asymmetric loss functions.

\begin{Definition}[Asymmetric Condition]
On the given weights $w_1, \dots, w_K \geq 0$, where $\exists t \in [K]$, s.t., $w_t > \max_{i \neq t} w_i$, a loss function $L$ is called asymmetric if $L$ satisfies
\begin{equation}
\arg\min_{f(\rvx)} \sum_{k=1}^K w_k L(f(\rvx),k) = \arg\min_{f(\rvx)} L(f(\rvx),t),
\end{equation}
where we always have $\arg\min_{f(\rvx)} L(f(\rvx),t) = \rve_t$.
\end{Definition}

\citet{ALFs_ICML, ALF_PAMI} proved that asymmetric loss functions are
noise-tolerant for clean-label-dominant noise, i.e., $1-\eta_\rvx > \max_{k \neq y} \eta_{\rvx, k}$,  $\forall \rvx$. However, existing asymmetric loss functions, such as Asymmetric Generalized Cross Entropy (AGCE) \cite{ALFs_ICML, ALF_PAMI}, are all active losses. This is because implementing the asymmetric condition in passive losses remains a challenging problem.

\vspace{4pt}
\noindent\textbf{Irreplaceable of NCE/NFL.} Although no passive asymmetric loss has been designed, can we replace the active NCE/NFL in the APL framework with an active asymmetric loss? To further explore this question, we conducted a series of experiments using active AGCE combined with passive MAE, as shown in Table~\ref{tab: agce+mae}. The results indicate that although AGCE+MAE adheres to the APL framework, it fails to achieve the desired effect. This suggests that simply replacing NCE with an asymmetric loss function within the APL framework does not lead to strong performance.
Currently, all robust loss functions based on the APL framework rely on NCE or its variant, NFL, as active losses, highlighting their crucial role in implementing the APL framework. Therefore, the key challenge is to design an effective passive asymmetric loss function that can be effectively integrated with NCE/NFL to further enhance the APL framework.
\begin{table}[!t]
\setlength{\tabcolsep}{4.5pt}
\fontsize{9.2pt}{11.5pt}\selectfont
\centering
\caption{Last epoch test accuracies (\%) of different methods on CIFAR-10 with symmetric ($\eta \in [0.4, 0.8]$) and asymmetric ($\eta \in [0.2, 0.4]$) label noise.  The results "mean$\pm$std" are reported over 3 random trials and the best results are in \textbf{bold}. $\dag$ RCE actually equals a scaled MAE \cite{SCE}. In order to be consistent with the original APL paper \cite{NCE}, we still write RCE here.}
\vskip-5pt
\label{tab: agce+mae}
\begin{tabular}{c|cc|cc}
\toprule
\multirow{2}{*}{\textbf{CIFAR-10}} & \multicolumn{2}{c|}{\textbf{Symmetric}}    & \multicolumn{2}{c}{\textbf{Asymmetric}}   \\
                                   & 0.4                 & 0.8                 & 0.2                 & 0.4                 \\
\midrule
MAE                                & 82.03\tiny $\pm$3.63          & 44.45\tiny $\pm$6.49          & 77.20\tiny $\pm$4.45          & 57.86\tiny $\pm$1.23          \\
NCE                                & 69.37\tiny $\pm$0.22          & 41.20\tiny $\pm$1.25          & 72.20\tiny $\pm$0.38          & 65.33\tiny $\pm$0.40          \\
AGCE                               & 83.39\tiny $\pm$0.17          & 44.42\tiny $\pm$0.74          & 86.67\tiny $\pm$0.14          & 60.91\tiny $\pm$0.20          \\
AGCE+MAE                           & 85.25\tiny $\pm$0.12          & 44.61\tiny $\pm$5.72          & 78.28\tiny $\pm$4.67          & 57.80\tiny $\pm$2.53          \\
NCE+RCE$^\dag$                            & \textbf{85.89\tiny $\pm$0.31} & \textbf{54.99\tiny $\pm$2.13} & \textbf{88.62\tiny $\pm$0.29} & \textbf{77.94\tiny $\pm$0.21} \\
\bottomrule
\end{tabular}
\end{table}

\subsection{Joint Asymmetric Loss}
In this paper, we extend the asymmetric loss function to a more complex passive loss scenario and propose the Asymmetric Mean Square Error (AMSE), a new asymmetric and passive loss function. Then, we embed the proposed AMSE into the APL framework to build a better performance framework, which we call Joint Asymmetric Loss (JAL). 

First, we introduce the proposed AMSE.

\vspace{4pt}
\noindent\textbf{Asymmetric Mean Square Error (AMSE):}
\begin{align}
    L_\text{AMSE}(f(\rvx), y) &= \frac{1}{K} \|a\cdot \rve_y - f(\rvx) \|_2^2 \notag\\
    &= \sum_{k=1}^K\frac{1}{K}|a\cdot e_k - f(\rvx)_k|^2, 
\end{align}
where $a \ge 1$ is a hyperparameter. AMSE is an extension of the MSE loss. 
If $a = 1$, this is the vanilla MSE loss. 

In the following, we build the sufficient and necessary condition for AMSE to realize the asymmetric condition.

\begin{Theorem}
\label{Theorem: amse}
On the given weights $w_1, \ldots, w_K$, where $w_m > w_n$, and $w_n = \max_{i \neq m} w_i$. The loss function 
$
L(f(\rvx), y) = \frac{1}{K} \| a \cdot \rve_y - f(\rvx) \|^q_q = \sum_{k=1}^K\frac{1}{K}|a\cdot e_k - f(\rvx)_k|^q
$
, where $q > 0$ and $a \geq 1$ are parameters, is asymmetric if and only if
$\frac{w_m}{w_n} \geq \frac{a^{q-1} + \sum_{i \neq m} \frac{w_i}{w_n}}{(a - 1)^{q-1}} \cdot \mathbb{I}(q > 1) + \mathbb{I}(q \leq 1)$.
\end{Theorem}

{\footnotesize
\noindent\textit{Proof.} 
For the sake of brevity, we abbreviate $f(\rvx)_k$ as $f_k$ in the proof. 

If $L(f(\mathbf{x}), k)$ is asymmetric, for $w_m > w_n \geq 0$, we have $\sum_{k=1}^K w_k L(f(\mathbf{x}), k) \geq \sum_{k=1}^K w_k L(f'(\mathbf{x}), k) \geq \sum_{k=1}^K w_k L(\mathbf{e}_m, k)$ always holds,  where $f'_i=f_i$ for $i = m, n$ and $f'_i = 0$ for $i \neq m, n$.
That is

$
w_m [(a - f_m)^q + f_n^q] + w_n [(a - f_n)^q + f_m^q] 
+ \sum_{i \neq m, n} w_i (a^q + f_m^q + f_n^q) \geq w_m (a - 1)^q + w_n (a^q + 1) + \sum_{i \neq m, n} w_i (a^q + 1).
$

For $w_n=0$, the The inequality is trivial.

For $w_n>0$, we have $\frac{w_m}{w_n}\ge$
\begin{equation*}
\begin{aligned}
    & \sup_{\substack{f_m,f_n\ge0\\ f_m+f_n=1}}\frac{a^q+1-(a-f_n)^q-f_m^q+{\displaystyle\sum_{i\neq m,n}}\frac{w_i}{w_n}(1-f_m^q-f_n^q)}{(a-f_m)^q+f_n^q-(a-1)^q} =\\
    & \sup_{0\le x\le 1}\frac{a^q+1-(a-1+x)^q-x^q +{\displaystyle\sum_{i\neq m,n}}\frac{w_i}{w_n}[1-x^q-(1-x)^q]}{(a-x)^q+(1-x)^q-(a-1)^q}\\
    & \triangleq\sup_{0\le x\le 1} h(x).
\end{aligned}
\end{equation*}

For $0< q\le 1$ and $0\le x \le 1$, because $a^q\le(a-1+x)^q+(1-x)^q$ and $1+(a-1)^q\le x^q+(a-x)^q$, we have $\frac{a^q+1-(a-1+x)^q-x^q }{(a-x)^q+(1-x)^q-(a-1)^q} \le 1$. Since $\sum_{i\neq m,n}\frac{w_i}{w_n}[1-x^q-(1-x)^q]\le0$, we have $\displaystyle\sup_{0\le x\le 1} h(x)=1$.

For $q> 1$, we have that $\sup_{0\le x\le 1} h(x)$ is equal to
\begin{equation*}
\begin{aligned}
    &\sup_{x\le \xi\le 1}\frac{(a-1+\xi)^{q-1}+\xi^{q-1}+\sum_{i\neq m,n}\frac{w_i}{w_n}[\xi^{q-1}-(1-\xi)^{q-1}]}{(a-\xi)^{q-1}+(1-\xi)^{q-1}}\\
    & \triangleq\sup_{x\le \xi\le 1} \rho(\xi) =\lim_{\xi\rightarrow 1} \rho(\xi)=\frac{a^{q-1}+\sum_{i\neq m}\frac{w_i}{w_n}}{(a-1)^{q-1}},
\end{aligned}
\end{equation*}
where the first line follows from Cauchy's Mean Value Theorem.

On the other hand, if $\frac{w_m}{w_n} \geq \frac{a^{q-1} + \sum_{i \neq m} \frac{w_i}{w_n}}{(a - 1)^{q-1}} \cdot \mathbb{I}(q > 1) + \mathbb{I}(q \leq 1)$. We reset $f'_m = f_m + f_n, f'_n = 0$, and $f'_i = f_i$ for $i \neq m, n$. We abbreviate $f_m+f_k$ as $f_{m\&k}$ for concision . Then for any $k \neq m$, we have 
$$
\begin{aligned}
&\frac{w_m}{w_k} \geq \frac{a^{q-1} + \sum_{i \neq m} \frac{w_i}{w_k}}{(a - 1)^{q-1}} \cdot \mathbb{I}(q > 1) + \mathbb{I}(q \leq 1) \Leftrightarrow \frac{w_m}{w_k} \geq \sup_{f_m, f_k \geq 0 \atop f_{m\&k} \le 1}\\
& \frac{  a^q + (f_{m\&k})^q - (a - f_k)^q - f_m^q + {\displaystyle\sum_{i \neq m, k}} \frac{w_i}{w_k} [(f_{m\&k})^q - f_m^q - f_k^q] }
{ (a - f_m)^q + f_k^q - (a - f_{m\&k})^q} \\
&\Rightarrow\sum_{k=1}^K w_k L(f(\rvx), k) \geq \sum_{k=1}^K w_k L(f'(\rvx), k),
\end{aligned}
$$
According to Lemma~1 in \cite{ALFs_ICML}, $L$ is asymmetric. \textit{End Proof.}
}

As shown in  Theorem~\ref{Theorem: amse}, we consider not only the case where $q = 2 $, but also other cases. 
To maintain consistency with MSE and simplify the loss function, we only use $q=2$ in the main paper. The analysis of different values of $q$ can be found in the supplementary materials.
Theorem~\ref{Theorem: amse} demonstrates that by adjusting a parameter $a$, 
AMSE, which is a passive loss, can satisfy the asymmetric condition and subsequently become noise-tolerant. For example, considering a 10-class dataset with 0.8 symmetric noise, we require  $\frac{w_m}{w_n} = \frac{0.2}{0.8/9}\ge \frac{a+9}{a-1}$, i.e., $a\ge 9$. 

\vspace{4pt}
\noindent\textbf{Parameter and Performance Analysis for AMSE.}
To demonstrate the superiority of the proposed AMSE, we compare it with the latest state-of-the-art passive loss, NNCE \cite{ANL}, on CIFAR-10.
Our analysis suggests that for CIFAR-10 with 0.8 symmetric noise, $a$ should be $\geq 9$. Therefore, we selected $a \in [10, 20, 30, 40]$ for our experiments, as shown in Figure~\ref{fig: acc line}.  
As illustrated, larger values of $a$ impose tighter constraints, making $a = 20, 30, 40$ more robust than $a = 10$ under 0.8 symmetric noise. However, excessively strict constraints may reduce the model's fitting ability, particularly under asymmetric noise. Therefore, selecting a moderate $a$ is recommended to achieve both robust and sufficient learning.
Overall, AMSE significantly outperforms NNCE in both symmetric and asymmetric noise, further demonstrating its effectiveness.
\begin{figure}[!t]
    \centering

    \begin{subfigure}{0.235\textwidth}  
        \centering
        \includegraphics[width=\textwidth]{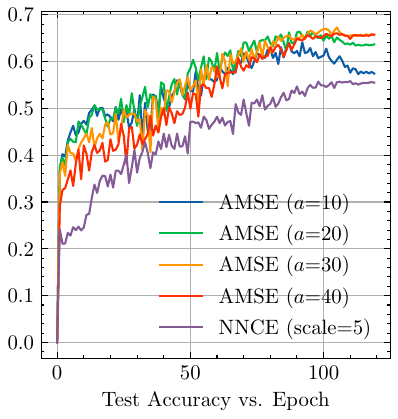}
        \caption{Symmetric 0.8}
        \label{fig: acc clean}
    \end{subfigure}
    \hfill
    \begin{subfigure}{0.235\textwidth}
        \centering
        \includegraphics[width=\textwidth]{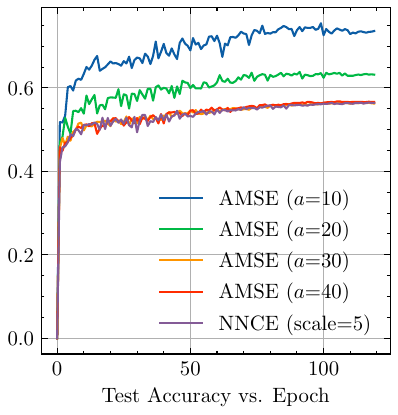}
        \caption{Asymmetric 0.4}
        \label{fig: acc 0.8}
    \end{subfigure}

    \vskip-5pt
    \caption{Test accuracies on CIFAR-10 with 0.8 symmetric and 0.4 asymmetric noise.}
    \label{fig: acc line}
\end{figure}

\begin{table}[!t]
\setlength{\tabcolsep}{4.4pt}
\centering
\caption{Last epoch test accuracies (\%) of different methods on CIFAR-10 with symmetric ($\eta \in [0.4, 0.8]$) and asymmetric ($\eta \in [0.2, 0.4]$) label noise.  The results "mean$\pm$std" are reported over 3 random trials and the best results are  in \textbf{bold}. }
\vskip-5pt
\label{tab: ablation}
\begin{tabular}{c|cc|cc}
\toprule
\multirow{2}{*}{\textbf{CIFAR-10}} & \multicolumn{2}{c|}{\textbf{Symmetric}}    & \multicolumn{2}{c}{\textbf{Asymmetric}}   \\
                                   & 0.4                 & 0.8                 & 0.2                 & 0.4                 \\
\midrule
NCE                                & 69.37\tiny $\pm$0.22          & 41.20\tiny $\pm$1.25          & 72.20\tiny $\pm$0.38          & 65.33\tiny $\pm$0.40          \\
AMSE                               & \textbf{87.54\tiny $\pm$0.26} & 64.97\tiny $\pm$0.87          & 83.88\tiny $\pm$5.07          & 58.07\tiny $\pm$2.21          \\
JAL-CE                           & 87.53\tiny $\pm$0.10          & \textbf{65.43\tiny $\pm$0.99} & \textbf{89.11\tiny $\pm$0.38} & \textbf{79.54\tiny $\pm$0.34} \\
\bottomrule
\end{tabular}
\end{table}

\begin{table*}[!t]
\centering
\setlength{\tabcolsep}{5.5pt}
\caption{Last epoch test accuracies (\%) of different methods on CIFAR-10 and CIFAR-100 with clean, symmetric ($\eta \in [0.2, 0.4, 0.6, 0.8]$), and asymmetric ($\eta \in [0.1, 0.2, 0.3, 0.4]$) label noise.  The results (mean$\pm$std) are reported over 3 random trials and the top-2 best results are in \textbf{bold}.}
\vskip-5pt
\label{tab: symm and asymm}
\begin{tabular}{c|c|cccc|cccc}
\toprule
\multirow{2}{*}{\textbf{CIFAR-10}}  & \multirow{2}{*}{\textbf{Clean}} & \multicolumn{4}{c|}{\textbf{Symmetric}}                                                & \multicolumn{4}{c}{\textbf{Asymmetric}}                                               \\
                                    &                                 & 0.2                 & 0.4                 & 0.6                 & 0.8                 & 0.1                 & 0.2                 & 0.3                 & 0.4                 \\
\midrule
CE                                  & 90.50\tiny $\pm$0.22                      & 75.21\tiny $\pm$0.39          & 58.05\tiny $\pm$0.53          & 38.80\tiny $\pm$0.45          & 19.74\tiny $\pm$0.40          & 86.85\tiny $\pm$0.15          & 83.05\tiny $\pm$0.35          & 78.37\tiny $\pm$0.61          & 73.85\tiny $\pm$0.07          \\
FL                                  & 89.70\tiny $\pm$0.24                      & 74.50\tiny $\pm$0.18          & 58.23\tiny $\pm$0.40          & 38.69\tiny $\pm$0.06          & 19.47\tiny $\pm$0.74          & 86.64\tiny $\pm$0.12          & 83.08\tiny $\pm$0.07          & 79.34\tiny $\pm$0.30          & 74.68\tiny $\pm$0.31          \\
GCE                                 & 89.36\tiny $\pm$0.19                      & 89.36\tiny $\pm$0.19          & 82.19\tiny $\pm$0.84          & 68.01\tiny $\pm$0.40          & 46.61\tiny $\pm$0.39          & 88.41\tiny $\pm$0.20          & 85.72\tiny $\pm$0.22          & 79.49\tiny $\pm$0.20          & 73.36\tiny $\pm$0.53          \\
SCE                                 & 91.51\tiny $\pm$0.24                      & 87.65\tiny $\pm$0.36          & 79.73\tiny $\pm$0.29          & 61.79\tiny $\pm$0.72          & 28.01\tiny $\pm$0.92          & 89.54\tiny $\pm$0.33          & 85.94\tiny $\pm$0.38          & 80.50\tiny $\pm$0.09          & 74.33\tiny $\pm$0.56          \\
NCE                                 & 75.48\tiny $\pm$0.37                      & 73.22\tiny $\pm$0.35          & 69.37\tiny $\pm$0.22          & 62.47\tiny $\pm$0.85          & 41.20\tiny $\pm$1.25          & 74.11\tiny $\pm$0.24          & 72.20\tiny $\pm$0.38          & 70.14\tiny $\pm$0.27          & 65.33\tiny $\pm$0.40          \\
NCE+RCE                             & 90.80\tiny $\pm$0.06                      & 88.93\tiny $\pm$0.04          & 85.89\tiny $\pm$0.31          & 79.89\tiny $\pm$0.25          & 54.99\tiny $\pm$2.13          & 90.04\tiny $\pm$0.17          & 88.62\tiny $\pm$0.29          & 85.07\tiny $\pm$0.27          & 77.94\tiny $\pm$0.21          \\
NCE+AUL                             & 91.17\tiny $\pm$0.18                      & 89.00\tiny $\pm$0.58          & 86.05\tiny $\pm$0.30          & 79.22\tiny $\pm$0.22          & 56.24\tiny $\pm$0.94          & 90.06\tiny $\pm$0.16          & 88.19\tiny $\pm$0.07          & 84.83\tiny $\pm$0.47          & 77.60\tiny $\pm$0.16          \\
NCE+AGCE                             & 91.01\tiny $\pm$0.20                      & 88.91\tiny $\pm$0.38          & 86.16\tiny $\pm$0.38          & 79.93\tiny $\pm$0.33          & 43.82\tiny $\pm$1.91          & 90.29\tiny $\pm$0.05          & 88.49\tiny $\pm$0.28          & 85.21\tiny $\pm$0.59          & 78.47\tiny $\pm$1.05          \\
CE+LC                               & 90.09\tiny $\pm$0.13                      & 83.87\tiny $\pm$0.27          & 70.36\tiny $\pm$0.23          & 46.53\tiny $\pm$0.29          & 19.74\tiny $\pm$1.77          & 87.74\tiny $\pm$0.23          & 83.16\tiny $\pm$0.33          & 78.48\tiny $\pm$0.25          & 73.32\tiny $\pm$0.78          \\
ANL-CE                              & 91.74\tiny $\pm$0.18                      & 89.68\tiny $\pm$0.29          & 87.16\tiny $\pm$0.16          & 81.28\tiny $\pm$0.63          & 62.28\tiny $\pm$1.10          & 90.66\tiny $\pm$0.16          & 89.09\tiny $\pm$0.21          & 85.49\tiny $\pm$0.49          & 77.99\tiny $\pm$0.40          \\
ANL-FL                              & 91.58\tiny $\pm$0.19                      & 89.93\tiny $\pm$0.03          & 86.94\tiny $\pm$0.03          & 81.10\tiny $\pm$0.30          & 61.89\tiny $\pm$2.25          & \textbf{90.72\tiny $\pm$0.20} & \textbf{89.29\tiny $\pm$0.02} & 85.80\tiny $\pm$0.38          & 77.89\tiny $\pm$0.28          \\
LT-APL	& -	& 89.42\tiny $\pm$0.13	& 86.82\tiny $\pm$0.18&	80.93\tiny $\pm$0.30&	40.87\tiny $\pm$1.57&	- &	89.28\tiny $\pm$0.24&	86.29\tiny $\pm$0.36	&\textbf{79.99\tiny $\pm$0.58} \\

\midrule
\textbf{JAL-CE}                     & 91.63\tiny $\pm$0.21                      & \textbf{89.95\tiny $\pm$0.22} & \textbf{87.53\tiny $\pm$0.10} & \textbf{82.03\tiny $\pm$0.18} & \textbf{65.43\tiny $\pm$0.99} & 90.70\tiny $\pm$0.21          & 89.11\tiny $\pm$0.38          & \textbf{86.38\tiny $\pm$0.14} & \textbf{79.54\tiny $\pm$0.34} \\
\textbf{JAL-FL}                     & 91.56\tiny $\pm$0.25                      & \textbf{89.99\tiny $\pm$0.11} & \textbf{87.43\tiny $\pm$0.29} & \textbf{82.09\tiny $\pm$0.08} & \textbf{64.84\tiny $\pm$1.13} & \textbf{90.77\tiny $\pm$0.16} & \textbf{89.36\tiny $\pm$0.27} & \textbf{86.18\tiny $\pm$0.04} & 79.51\tiny $\pm$0.06 \\
\midrule\midrule
\multirow{2}{*}{\textbf{CIFAR-100}} & \multirow{2}{*}{\textbf{Clean}} & \multicolumn{4}{c|}{\textbf{Symmetric}}                                                & \multicolumn{4}{c}{\textbf{Asymmetric}}                                               \\
                                    &                                 & 0.2                 & 0.4                 & 0.6                 & 0.8                 & 0.1                 & 0.2                 & 0.3                 & 0.4                 \\
\midrule
CE                                  & 70.93\tiny $\pm$0.77                      & 56.47\tiny $\pm$1.34          & 39.68\tiny $\pm$0.77          & 22.64\tiny $\pm$0.53          & 7.82\tiny $\pm$0.33           & 64.14\tiny $\pm$1.01          & 58.67\tiny $\pm$0.45          & 50.44\tiny $\pm$1.16          & 41.51\tiny $\pm$0.12          \\
FL                                  & 70.58\tiny $\pm$0.34                      & 56.32\tiny $\pm$1.43          & 40.83\tiny $\pm$0.52          & 22.44\tiny $\pm$0.54          & 7.68\tiny $\pm$0.37           & 65.00\tiny $\pm$0.46          & 58.12\tiny $\pm$0.44          & 51.16\tiny $\pm$1.32          & 41.46\tiny $\pm$0.38          \\
GCE                                  & 61.73\tiny $\pm$1.30                      & 60.58\tiny $\pm$2.51          & 57.35\tiny $\pm$0.91          & 46.15\tiny $\pm$1.10          & 20.33\tiny $\pm$0.31          & 62.01\tiny $\pm$1.11          & 59.19\tiny $\pm$1.36          & 53.35\tiny $\pm$0.65          & 40.92\tiny $\pm$0.21          \\
SCE                                 & 70.57\tiny $\pm$0.93                      & 55.50\tiny $\pm$0.35          & 40.13\tiny $\pm$1.48          & 22.23\tiny $\pm$1.29          & 7.84\tiny $\pm$0.56           & 64.51\tiny $\pm$0.45          & 57.84\tiny $\pm$0.57          & 49.66\tiny $\pm$0.48          & 41.58\tiny $\pm$0.87          \\
NCE                                 & 29.95\tiny $\pm$0.56                      & 25.43\tiny $\pm$0.91          & 20.26\tiny $\pm$0.25          & 14.66\tiny $\pm$1.04          & 8.82\tiny $\pm$0.47           & 27.16\tiny $\pm$1.01          & 26.67\tiny $\pm$0.73          & 23.83\tiny $\pm$0.29          & 20.83\tiny $\pm$1.08          \\
NCE+RCE                             & 68.07\tiny $\pm$0.70                      & 64.57\tiny $\pm$0.16          & 58.48\tiny $\pm$0.51          & 46.73\tiny $\pm$1.00          & 26.94\tiny $\pm$1.29          & 66.74\tiny $\pm$0.30          & 62.82\tiny $\pm$0.57          & 55.86\tiny $\pm$0.40          & 41.50\tiny $\pm$0.39          \\
NCE+AUL                             & 69.95\tiny $\pm$0.33                      & 65.45\tiny $\pm$0.49          & 56.37\tiny $\pm$0.12          & 38.68\tiny $\pm$0.75          & 12.95\tiny $\pm$0.37          & 66.41\tiny $\pm$0.15          & 57.39\tiny $\pm$0.34          & 48.20\tiny $\pm$0.19          & 38.41\tiny $\pm$0.52          \\
NCE+AGCE                            & 69.05\tiny $\pm$0.36                      & 65.61\tiny $\pm$0.27          & 59.40\tiny $\pm$0.34          & 47.66\tiny $\pm$0.49          & 26.14\tiny $\pm$0.01          & 66.96\tiny $\pm$0.45          & 64.08\tiny $\pm$0.44          & 57.17\tiny $\pm$0.33          & 44.62\tiny $\pm$1.04          \\
CE+LC                               & 71.80\tiny $\pm$0.34                      & 56.26\tiny $\pm$0.09          & 37.36\tiny $\pm$0.49          & 17.46\tiny $\pm$0.62          & 6.32\tiny $\pm$0.16           & 63.51\tiny $\pm$0.27          & 56.19\tiny $\pm$0.30          & 48.07\tiny $\pm$0.38          & 39.64\tiny $\pm$0.14          \\
ANL-CE                              & 70.26\tiny $\pm$0.15                      & 66.93\tiny $\pm$0.09          & 61.58\tiny $\pm$0.33          & 52.09\tiny $\pm$0.58          & \textbf{28.01\tiny $\pm$1.06}          & 68.60\tiny $\pm$0.41          & 65.96\tiny $\pm$0.18          & 60.57\tiny $\pm$0.07          & 45.73\tiny $\pm$0.74          \\
ANL-FL                              & 70.11\tiny $\pm$0.27                      & 67.03\tiny $\pm$0.46          & 61.89\tiny $\pm$0.25          & 51.58\tiny $\pm$0.33          & \textbf{28.81\tiny $\pm$0.74}          & 68.67\tiny $\pm$0.21          & 66.12\tiny $\pm$0.39          & 60.03\tiny $\pm$0.48          & 46.20\tiny $\pm$0.45          \\
LT-APL	& - &	63.29\tiny $\pm$0.49&	54.70\tiny $\pm$1.73&	40.52\tiny $\pm$1.65&	22.63\tiny $\pm$0.78&	-&	62.59\tiny $\pm$1.31&	56.90\tiny $\pm$1.29&	44.05\tiny $\pm$1.32\\

\midrule
\textbf{JAL-CE}                     & 70.60\tiny $\pm$0.09                      & \textbf{68.25\tiny $\pm$0.39} & \textbf{64.11\tiny $\pm$0.55} & \textbf{56.73\tiny $\pm$0.65} & 22.80\tiny $\pm$2.11 & \textbf{69.29\tiny $\pm$0.42} & \textbf{67.90\tiny $\pm$0.59} & \textbf{64.90\tiny $\pm$0.27} & \textbf{56.17\tiny $\pm$0.32} \\
\textbf{JAL-FL}                     & 70.66\tiny $\pm$0.37                      & \textbf{68.33\tiny $\pm$0.34} & \textbf{64.55\tiny $\pm$0.61} & \textbf{56.44\tiny $\pm$0.22} & 23.11\tiny $\pm$2.28  & \textbf{69.25\tiny $\pm$0.21} & \textbf{67.63\tiny $\pm$0.50} & \textbf{65.18\tiny $\pm$0.26} &  \textbf{56.26\tiny $\pm$0.05} \\
\bottomrule
\end{tabular}
\end{table*}

\vspace{4pt}
\noindent\textbf{Joint Asymmetric Loss.}
We now integrate the proposed AMSE into the APL framework to enhance its performance, resulting in a novel approach called Joint Asymmetric Loss (JAL). Specifically, we introduce two joint asymmetric losses, as described in the following.

Base on Cross Entropy (CE), we have JAL-CE:
\begin{equation}
  L_\text{JAL-CE} = \alpha\cdot L_\text{NCE} + \beta \cdot L_\text{AMSE}  .
\end{equation}

Base on Focal Loss (FL), we have JAL-FL:
\begin{equation}
  L_\text{JAL-FL} = \alpha\cdot L_\text{NFL} + \beta \cdot L_\text{AMSE}  .
\end{equation}

We can easily prove that JAL remains noise-tolerant.  
\citet{ALFs_ICML} demonstrated that symmetric loss functions are completely asymmetric and that the combination of asymmetric loss functions remains asymmetric. Since NCE/NFL are symmetric (and therefore also asymmetric), and we have already proven that AMSE is asymmetric, it follows that JAL is also asymmetric and thus noise-tolerant.


\noindent\textbf{Robust and Sufficient learning of JAL.} To evaluate the effectiveness of our proposed JAL framework in improving performance, we conducted ablation experiments on CIFAR-10 using NCE, AMSE ($a=30$), and JAL-CE ($\alpha=1, \beta=1, a=30$), as shown in Table~\ref{tab: ablation}.  
The results indicate that under symmetric noise, JAL-CE performs similarly to AMSE, with both achieving strong performance. This highlight the effectiveness of the the AMSE component. In addition, JAL framework can effectively alleviate parameter sensitivity to noise rates and types. Under asymmetric noise, NCE and AMSE exhibit signs of underfitting, whereas JAL-CE maintains a strong fitting ability.  These findings demonstrate that 
JAL offers both robustness and superior fitting ability in label noise scenarios.

\section{Experiments}
In this section, we provide extensive experiments to evaluate the effectiveness of our method on various datasets, including CIFAR-10/CIFAR-100 \cite{cifar}, CIFAR-10N/CIFAR-100N \cite{cifar-n}, WebVision \cite{webvision}, ILSVRC12 \cite{imagenet}, and Clothing1M \cite{clothing1m}. 
Detailed experiment settings can be found in the supplementary materials.

\begin{table*}[!t]
\centering
\caption{Last epoch test accuracies (\%) of different methods on CIFAR-10 and CIFAR-100 with instance-dependent noise (IDN) ($\eta \in [0.2, 0.4, 0.6]$).  The results "mean$\pm$std" are reported over 3 random trials and the top-2 best results are in \textbf{bold}.}
\vskip-5pt
\label{tab: idn}
\begin{tabular}{c|ccc|ccc}
\toprule
\multirow{2}{*}{\textbf{Loss}} & \multicolumn{3}{c|}{\textbf{CIFAR-10 IDN}}                       & \multicolumn{3}{c}{\textbf{CIFAR-100 IDN}}                      \\
                               & 0.2                 & 0.4                 & 0.6                 & 0.2                 & 0.4                 & 0.6                 \\
\toprule
CE                             & 75.38\tiny $\pm$0.19          & 57.63\tiny $\pm$0.27          & 37.97\tiny $\pm$0.36          & 57.02\tiny $\pm$0.54          & 40.91\tiny $\pm$2.05          & 24.49\tiny $\pm$0.86          \\
GCE                            & 86.66\tiny $\pm$0.14          & 79.99\tiny $\pm$0.23          & 51.90\tiny $\pm$0.13          & 61.43\tiny $\pm$2.24          & 57.07\tiny $\pm$1.04          & 42.40\tiny $\pm$0.52          \\
SCE                            & 86.65\tiny $\pm$0.27          & 74.54\tiny $\pm$0.34          & 49.83\tiny $\pm$0.40          & 56.32\tiny $\pm$0.27          & 39.82\tiny $\pm$1.43          & 23.19\tiny $\pm$0.87          \\
NCE+RCE                        & 89.06\tiny $\pm$0.26          & 85.11\tiny $\pm$0.28          & 71.27\tiny $\pm$0.66          & 64.33\tiny $\pm$0.46          & 57.53\tiny $\pm$0.84          & 40.36\tiny $\pm$0.35          \\
NCE+AGCE                       & 88.95\tiny $\pm$0.07          & 85.30\tiny $\pm$0.23          & 71.49\tiny $\pm$0.34          & 65.18\tiny $\pm$0.17          & 57.89\tiny $\pm$0.57          & 43.04\tiny $\pm$0.29          \\
CE+LC                          & 82.77\tiny $\pm$0.09          & 68.06\tiny $\pm$0.22          & 43.60\tiny $\pm$0.39          & 55.93\tiny $\pm$0.39          & 37.74\tiny $\pm$0.63          & 18.68\tiny $\pm$0.50          \\
ANL-CE                         & 89.71\tiny $\pm$0.35          & 85.74\tiny $\pm$0.15          & 69.83\tiny $\pm$0.38          & 66.89\tiny $\pm$0.53          & 60.88\tiny $\pm$0.35          & 48.12\tiny $\pm$0.48          \\
ANL-FL                         & 89.68\tiny $\pm$0.21          & 85.97\tiny $\pm$0.16          & 70.70\tiny $\pm$0.30          & 67.17\tiny $\pm$0.11          & 61.07\tiny $\pm$0.38          & 46.77\tiny $\pm$0.80          \\
\midrule
\textbf{JAL-CE}                & \textbf{90.01\tiny $\pm$0.12} & \textbf{86.46\tiny $\pm$0.15} & \textbf{75.62\tiny $\pm$0.18} & \textbf{67.51\tiny $\pm$0.29} & \textbf{63.24\tiny $\pm$0.16} & \textbf{51.69\tiny $\pm$0.68} \\
\textbf{JAL-FL}                & \textbf{89.90\tiny $\pm$0.14} & \textbf{86.78\tiny $\pm$0.17} & \textbf{75.02\tiny $\pm$0.48} & \textbf{67.77\tiny $\pm$0.38} & \textbf{63.56\tiny $\pm$0.18} & \textbf{51.69\tiny $\pm$0.59} \\
\bottomrule
\end{tabular}
\end{table*}
\begin{table}[!t]
\setlength{\tabcolsep}{3pt}
\fontsize{9.2pt}{11.5pt}\selectfont
\centering
\caption{Last epoch test accuracies (\%) of different methods on CIFAR-10N and CIFAR-100N human-annotated noise \cite{cifar-n}.  The results "mean$\pm$std" are reported over 3 random trials and the top-2 best results are in \textbf{bold}.}
\vskip-5pt
\label{tab: cifar_n}
\begin{tabular}{c|ccc|c}
\toprule
\multirow{2}{*}{\textbf{Loss}} & \multicolumn{3}{c|}{\textbf{CIFAR-10}}                          & \textbf{CIFAR-100} \\
                               & Aggregate           & Random 1            & Worst               & Noisy               \\
\midrule
CE                             & 85.09\tiny $\pm$0.30          & 79.09\tiny $\pm$0.28          & 61.43\tiny $\pm$0.52          & 48.63\tiny $\pm$0.53          \\
GCE                            & 87.39\tiny $\pm$0.09          & 85.98\tiny $\pm$0.42          & 77.77\tiny $\pm$0.59          & 50.97\tiny $\pm$0.60          \\
SCE                            & 88.48\tiny $\pm$0.26          & 85.65\tiny $\pm$0.30          & 73.65\tiny $\pm$0.29          & 48.52\tiny $\pm$0.11          \\
NCE+RCE                        & 89.17\tiny $\pm$0.28          & 87.62\tiny $\pm$0.34          & 79.74\tiny $\pm$0.09          & 54.27\tiny $\pm$0.09          \\
NCE+AGCE                     & 89.27\tiny $\pm$0.28          & 87.92\tiny $\pm$0.02          & 79.91\tiny $\pm$0.37          & 55.96\tiny $\pm$0.20          \\
CE+LC                          & 86.60\tiny $\pm$0.40          & 83.51\tiny $\pm$0.13          & 70.11\tiny $\pm$0.10          & 47.76\tiny $\pm$0.29          \\
ANL-CE                         & 89.66\tiny $\pm$0.12          & 88.68\tiny $\pm$0.13          & 80.23\tiny $\pm$0.28          & 56.37\tiny $\pm$0.42          \\
ANL-FL                         & 89.81\tiny $\pm$0.08          & 88.57\tiny $\pm$0.18          & 80.56\tiny $\pm$0.23          & 57.09\tiny $\pm$0.40          \\
\midrule
\textbf{JAL-CE}                & \textbf{89.94\tiny $\pm$0.20} & \textbf{88.85\tiny $\pm$0.23} & \textbf{81.33\tiny $\pm$0.34} & \textbf{59.54\tiny $\pm$0.12} \\
\textbf{JAL-FL}                & \textbf{90.06\tiny $\pm$0.22} & \textbf{88.71\tiny $\pm$0.30} & \textbf{81.25\tiny $\pm$0.10} & \textbf{59.38\tiny $\pm$0.24} \\
\bottomrule
\end{tabular}
\end{table}
\subsection{Evaluation on Benchmark Datasets} 
\noindent\textbf{Baselines.} 
We experiment with various state-of-the-art methods, including
Cross Entropy (CE); Focal Loss (FL) \cite{FL}, Generalized Cross Entropy (GCE) \citep{GCE}; Symmetric Cross Entropy (SCE) \citep{SCE};  
Active Passive Loss (APL) \citep{NCE}, including NCE and NCE+RCE \citep{NCE}; Asymmetric Loss Functions (ALFs) \citep{ALFs_ICML, ALF_PAMI}, including NCE+AUL and NCE+AGCE;  LogitClip (CE+LC) \cite{LC};
Active Negative Loss (ANL) \citep{ANL}, including ANL-CE and ANL-FL. Student loss (LT-APL) \citep{student_loss}. We follow the same experimental settings in \citep{NCE, ALFs_ICML, ANL}: An 8-layer CNN \cite{cnn} is used for CIFAR-10 and a ResNet-34 \citep{resnet} for CIFAR-100.



\vspace{4pt}
\noindent\textbf{Results.}
We evaluate the test accuracy of various methods under different types of label noise, including symmetric, asymmetric, instance-dependent, and human-annotated noise. 
The experimental results for symmetric and asymmetric noise are presented in Table~\ref{tab: symm and asymm}.
As shown, our proposed JAL-CE and JAL-FL demonstrate exceptional performance, consistently ranking among the top-2 in most scenarios. On CIFAR-10, under the most challenging 0.8 symmetric noise, JAL achieves an improvement in accuracy of about 3\%.
For the more complex CIFAR-100, our method significantly outperforms previous state-of-the-art methods in most cases. In particular, on CIFAR-100, our method improves accuracy by 10\% under 0.4 asymmetric noise.

The experimental results for instance-dependent noise (IDN) are presented in Table~\ref{tab: idn}. As can be seen, our JAL-CE and JAL-FL consistently achieve top-2 performance across all cases, with a 3\textasciitilde4\% increase in accuracy under 0.6 instance-dependent noise on both CIFAR-10 and CIFAR-100 compared to previous state-of-the-art methods, such as NCE+RCE, NCE+AGCE, and ANL.

Furthermore, we conduct experiments on human-annotated label noise using the CIFAR-10N and CIFAR-100N datasets \cite{cifar-n}, as shown in Table \ref{tab: cifar_n}. As can be seen, our method achieves top-2 performance across all human-annotated cases, especially for the most difficult CIFAR-10 worst and CIFAR-100 noisy cases, highlighting the excellent performance of our method in practical scenarios. 
These results demonstrate that our method significantly surpasses the latest benchmarks.
\begin{figure*}[!t]
    \centering
    \begin{subfigure}{0.235\textwidth}  
        \centering
        \includegraphics[width=\textwidth]{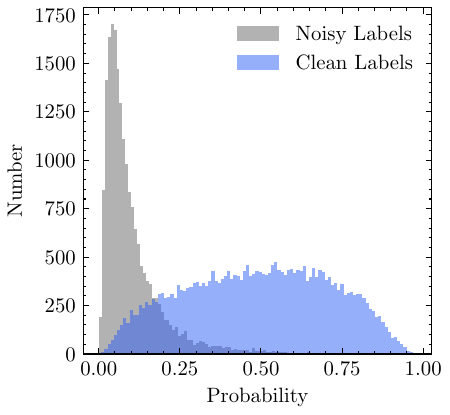}
        \caption{CE with 30 Epochs}
        \label{fig: hist_ce_ep=30}
    \end{subfigure}
    \hfill
    \begin{subfigure}{0.23\textwidth}
        \centering
        \includegraphics[width=\textwidth]{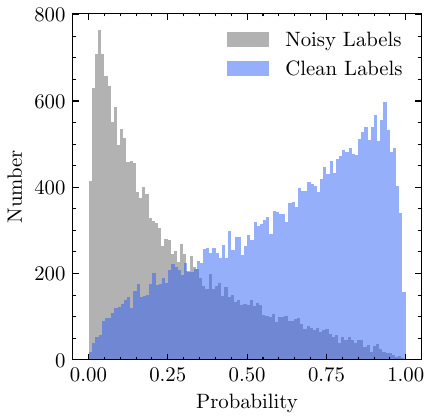}
        \caption{CE with 60 Epochs}
        \label{fig: hist_ce_ep=60}
    \end{subfigure}
    \hfill
    \begin{subfigure}{0.235\textwidth}
        \centering
        \includegraphics[width=\textwidth]{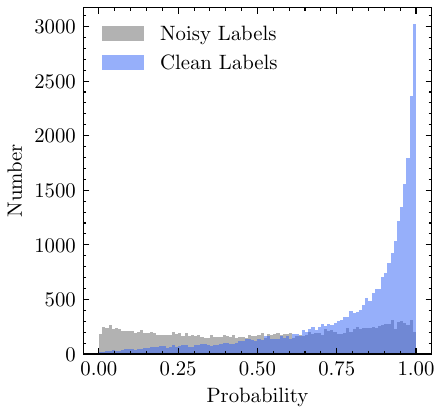}
        \caption{CE with 90 Epochs}
        \label{fig: hist_ce_ep=90}
    \end{subfigure}
    \hfill
    \begin{subfigure}{0.235\textwidth}
        \centering
        \includegraphics[width=\textwidth]{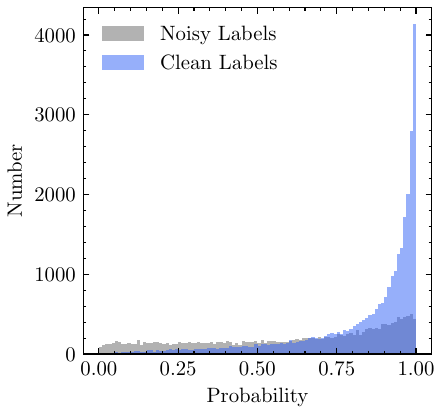}
        \caption{CE with 120 Epochs}
        \label{hist_ce_ep=120}
    \end{subfigure}
    \\
    \begin{subfigure}{0.235\textwidth}
        \centering
        \includegraphics[width=\textwidth]{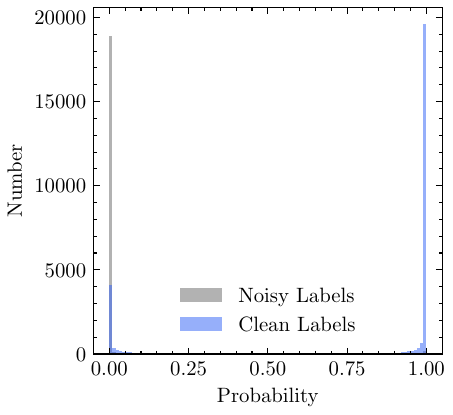}
        \caption{JAL-CE with 30 Epochs}
        \label{hist_ce_ep=30}
    \end{subfigure}
    \hfill
    \begin{subfigure}{0.235\textwidth}
        \centering
        \includegraphics[width=\textwidth]{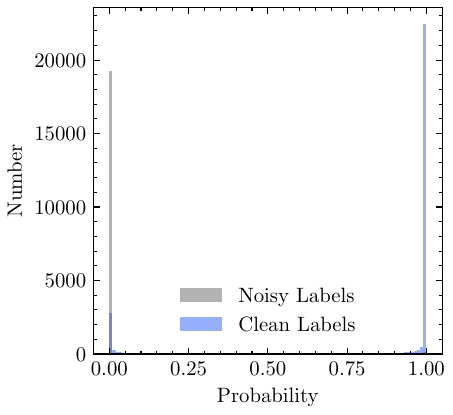}
        \caption{JAL-CE with 60 Epochs}
        \label{hist_ce_ep=60}
    \end{subfigure}
    \hfill
    \begin{subfigure}{0.235\textwidth}
        \centering
        \includegraphics[width=\textwidth]{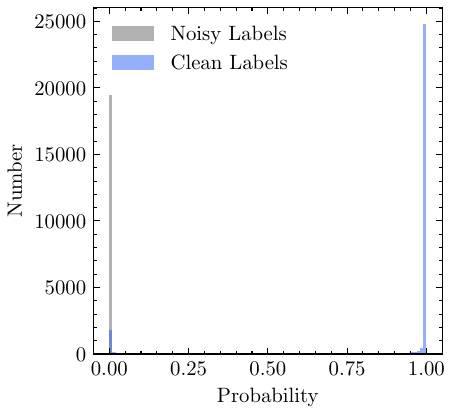}
        \caption{JAL-CE with 90 Epochs}
        \label{hist_ce_ep=90}
    \end{subfigure}
    \hfill
    \begin{subfigure}{0.235\textwidth}
        \centering
        \includegraphics[width=\textwidth]{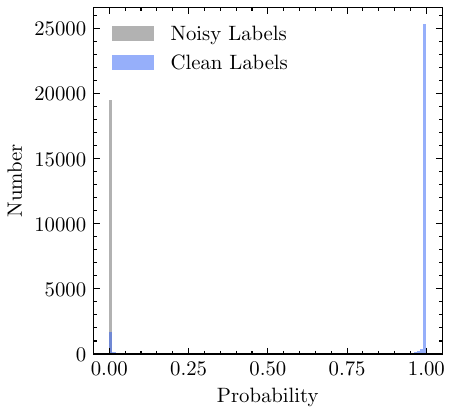}
        \caption{JAL-CE with 120 Epochs}
        \label{hist_ce_ep=120}
    \end{subfigure}

    \vskip-5pt
    \caption{Histograms of the distribution of samples with different prediction probabilities in the training set for CIFAR-10 with 0.4 symmetric noise.}
    \label{fig: hist}
\end{figure*}

\begin{table*}[!t]
\centering
\caption{Last epoch test accuracies (\%) of different methods on ILSVRC12, WebVision, and Clothing1M. 
The top-2 best results are in \textbf{bold}.}
\vskip-5pt
\label{tab: real_world}
\begin{tabular}{c|ccccccccc}
\toprule
\textbf{Loss}       & CE    & GCE   & SCE   & NCE+RCE & NCE+AGCE & ANL-CE & ANL-FL & \textbf{JAL-CE} & \textbf{JAL-FL} \\
\midrule
\textbf{WebVision}  & 66.28 & 61.84 & 65.16 & 66.96   & 67.16    & 67.36  & 67.76  & \textbf{69.84}  & \textbf{69.20}   \\
\midrule
\textbf{ILSVRC12}   & 60.68 & 60.32 & 61.00    & 63.96   & 64.36    & 65.60   & 64.84  & \textbf{66.64}  & \textbf{66.00}     \\
\midrule
\textbf{Clothing1M} & 67.93 & 68.46 & 67.71 & 69.24   & 67.90     & 69.75  & 69.90   & \textbf{70.31}  & \textbf{70.11}  \\
\bottomrule
\end{tabular}
\end{table*}
\vspace{4pt}
\noindent\textbf{Comparison with Previous Asymmetric Loss Functions.}
Previous asymmetric loss functions have also been combined with NCE to enhance performance, such as NCE+AUL and NCE+AGCE. However, since both NCE and earlier asymmetric loss functions act as active losses, they do not form the APL framework. As a result, only limited improvements can be gained, which explains why our JAL method outperforms previous asymmetric loss approaches.

\vspace{4pt}
\noindent\textbf{Histogram Visualization.}
To further assess the robustness of our JAL method compared to vanilla CE, we visualize the prediction probability distributions on the training set for models trained on CIFAR-10 with 0.4 symmetric noise, as illustrated in Figure~\ref{fig: hist}.
The results reveal that while CE initially fits clean labels in the early training stages, it progressively overfits to noisy labels as training continues. In contrast, JAL demonstrates superior robustness by predominantly focusing on clean labels while effectively avoiding fitting to noisy labels throughout all the training process.

\subsection{Evaluation on Real-World Datasets}

We perform experiments on large-scale real-world datasets, including WebVision \citep{webvision}, ILSVRC12 \citep{imagenet}, and Clothing1M \citep{clothing1m}. For WebVision, we follow the mini setting in \cite{mentornet} that takes the first 50 classes in the google image subset. We train a ResNet-50 \cite{resnet} and evaluate the trained network on the same 50 classes of ILSVRC12 and WebVision validation set.
For Clothing1M, we use a ResNet-50 pre-trained on ImageNet similar to \citep{clothing1m}. We train the model on the noisy training set with a million samples and subsequently evaluate it on the test set.

\vspace{4pt}
\noindent\textbf{Results.}
In Table \ref{tab: real_world}, we present the test accuracies achieved by various robust loss functions on ILSVRC12, WebVision, and Clothing1M. Notably, our JAL-CE and JAL-FL outperform other state-of-the-art methods, achieving the highest accuracy across all real-world datasets. These results highlight the robustness and effectiveness of JAL in practical applications.

\section{Conclusion}
In this paper, we expand the research of asymmetric loss functions, and realize a more complex passive asymmetric loss function. 
Specifically, we introduce the \textit{Asymmetric Mean Square Error} (AMSE), the first passive asymmetric loss function. We rigorously establish the necessary and sufficient condition for AMSE to satisfy the asymmetric condition.
By replacing the traditional passive symmetric loss in APL with our AMSE, we propose the \textit{Joint Asymmetric Loss} (JAL), a novel robust loss framework with better fitting ability.
Our theoretically guaranteed method has shown positive results in mitigating label noise. 
We hope AMSE and JAL will be useful with other methods and tasks that involve noise-tolerant learning.

\section*{Acknowledgements}
This work was supported in part by National Natural Science Foundation of China under Grants 632B2031 and 92270116, in part by National Key Research and Development Program of China under Grant 2023YFC2509100, and in part by HIT-XNJKKGYDJ2024014.

{
    \small
    \bibliographystyle{ieeenat_fullname}
    \bibliography{main}
}
\newpage
\appendix
\onecolumn
\section*{\centering\Large Joint Asymmetric Loss for Learning with Noisy Labels \\
Supplementary Materials}

\vspace{20pt}

\section{More Results}

\noindent\textbf{More Ablation Experiments about AMSE.} We present the ablation experiments for different $q$ and $a$ in Figure~\ref{fig: acc line}. As illustrated: 
1) For $q=1$, the asymmetric condition always holds. In this case, $a$ is a constant with zero gradient, making different choices of $a$ equivalent. 
The loss is difficult to optimize, similar to MAE.  
2) For $q=2$, the asymmetric condition holds when $a \ge 9$. For the gradient, we have $\frac{\partial L(f(\mathbf{x}),y)}{\partial f(\mathbf{x})_y} = -\frac{2}{K}(a - f(\mathbf{x})_y)$, and $a$ does not affect $\frac{\partial L(f(\mathbf{x}),y)}{\partial f(\mathbf{x})_{k \neq y}}$. As $a$ increases, the weight of high-confidence (clean) samples in the gradient increases, while the weight of low-confidence (noisy) samples decreases. This explains why a larger $a$ leads to better robustness.  
3) For $q=3$, the condition holds when $a \ge 4.73$. The performance of the loss is similar to $q=2$, but it is more sensitive to the hyperparameter, as higher powers amplify the loss error. Therefore, using $q=2$ is an appropriate choice.
\begin{figure}[!th]
    \centering

    \begin{subfigure}{0.235\textwidth}  
        \centering
        \includegraphics[width=\textwidth]{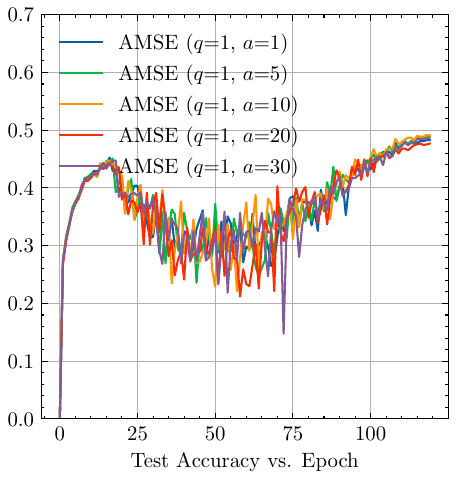}
        \caption{AMSE ($q=1$)}
        \label{fig: q=1}
    \end{subfigure}
    \hspace{20pt}
    \begin{subfigure}{0.235\textwidth}
        \centering
        \includegraphics[width=\textwidth]{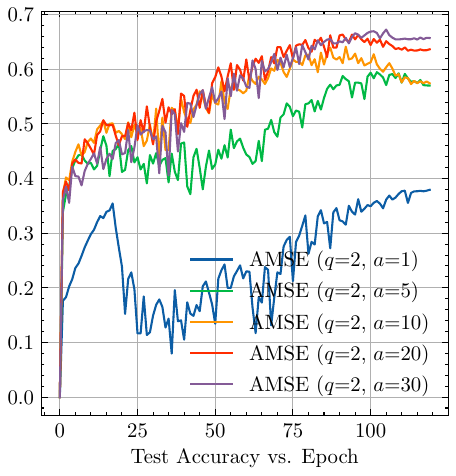}
        \caption{AMSE ($q=2$)}
        \label{fig: q=2}
    \end{subfigure}
    \hspace{20pt}
    \begin{subfigure}{0.235\textwidth}  
        \centering
        \includegraphics[width=\textwidth]{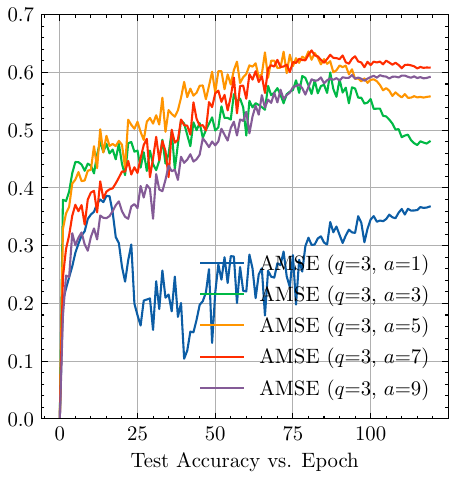}
        \caption{AMSE ($q=3$)}
        \label{fig: q=3}
    \end{subfigure}

    \vskip-5pt
    \caption{Ablation experiments for AMSE on CIFAR-10 with 0.8 symmetric noise.}
    \label{fig: acc line}
\end{figure}

\vspace{4pt}
\noindent\textbf{More Results for AGCE+MAE.} For the experiment for AGCE+MAE, we use the same $a=6, q=1.5$ in \cite{ALFs_ICML}, and search for $\alpha, \beta \in [1, 10]$. The complete results are presented in Table~\ref{tab: agce+mae all}, while the results for $\alpha=1, \beta=1$ are shown in the main paper.

\begin{table*}[!ht]
\centering
\caption{Last epoch test accuracies (\%) of different methods on CIFAR-10 with symmetric ($\eta \in [0.4, 0.8]$) and asymmetric ($\eta \in [0.2, 0.4]$) label noise.  The results "mean$\pm$std" are reported over 3 random trials and the best results are in \textbf{bold}. $\dag$ RCE actually equals a scaled MAE \cite{SCE}. In order to be consistent with the original APL paper \cite{NCE}, we still write RCE here.}
\vskip-5pt
\label{tab: agce+mae all}
\begin{tabular}{c|cc|cc}
\toprule
\multirow{2}{*}{\textbf{CIFAR-10}} & \multicolumn{2}{c|}{\textbf{Symmetric}}                        & \multicolumn{2}{c}{\textbf{Asymmetric}}   \\
                                   & 0.4                 & 0.8                                     & 0.2                 & 0.4                 \\
                                   \midrule
MAE                                & 82.03\tiny $\pm$3.63          & 44.45\tiny $\pm$6.49                              & 77.20\tiny $\pm$4.45          & 57.86\tiny $\pm$1.23          \\
NCE                                & 69.37\tiny $\pm$0.22          & 41.20\tiny $\pm$1.25                              & 72.20\tiny $\pm$0.38          & 65.33\tiny $\pm$0.40          \\
AGCE                               & 83.39\tiny $\pm$0.17          & 44.42\tiny $\pm$0.74                              & 86.67\tiny $\pm$0.14          & 60.91\tiny $\pm$0.20          \\
AGCE+MAE $(\alpha=1, \beta=1)$          & 85.25\tiny $\pm$0.12          & 44.61\tiny $\pm$5.72                              & 78.28\tiny $\pm$4.67          & 57.80\tiny $\pm$2.53          \\
AGCE+MAE $(\alpha=1, \beta=10)$         & 85.86\tiny $\pm$0.11          & 39.44\tiny $\pm$0.71                              & 77.64\tiny $\pm$3.75          & 56.50\tiny $\pm$0.41          \\
AGCE+MAE $(\alpha=10, \beta=1)$         & 85.71\tiny $\pm$0.29          & 23.36\tiny $\pm$2.85                              & 75.43\tiny $\pm$4.16          & 57.55\tiny $\pm$1.83          \\
AGCE+MAE $(\alpha=10, \beta=10)$        & 85.85\tiny $\pm$0.55          & 21.83\tiny $\pm$1.47                              & 78.92\tiny $\pm$4.59          & 56.49\tiny $\pm$0.50          \\
NCE+RCE$^\dag$        & \textbf{85.89\tiny $\pm$0.31} & \textbf{54.99\tiny $\pm$2.13} & \textbf{88.62\tiny $\pm$0.29} & \textbf{77.94\tiny $\pm$0.21} \\
\bottomrule
\end{tabular}
\end{table*}

\section{Experiments}

\subsection{Evaluation on Benchmark Datasets}
\noindent\textbf{Noise Generation.}
We follow the approach of the previous work \citep{ANL} to experiment with two types of synthetic label noise: symmetric noise and asymmetric noise. In the case of symmetric label noise, we intentionally corrupt the training labels by randomly flipping labels within each class to incorrect labels in other classes. As for asymmetric label noise, we flip the labels within a specific sets of classes: For CIFAR-10, the flips occur from TRUCK $\rightarrow$ AUTOMOBILE, BIRD $\rightarrow$ AIRPLANE, DEER $\rightarrow $ HORSE, and CAT $\leftrightarrow$ DOG. For CIFAR-100, the 100 classes are grouped into 20 super-classes, each containing 5 sub-classes, and we flip the labels within the same super-class into the next. For instance-dependent noise, we follow the approach in PDN \citep{IDN-PDN} for generating label noise.

\vspace{4pt}
\noindent\textbf{Experimental Setting.} 
We follow the experimental settings in \citep{NCE, ALFs_ICML,ANL}: An 8-layer CNN is used for CIFAR-10  and a ResNet-34 \citep{cnn, resnet} for CIFAR-100. The networks are trained for 120 and 200 epochs for CIFAR-10 and CIFAR-100  with batch size 128. We use the SGD optimizer with momentum 0.9 and L1 weight decay  $5\times10^{-5}$ and $5\times10^{-6}$ for CIFAR-10 and CIFAR-100.
The learning rate is set to 0.01 for CIFAR-10 and 0.1 for CIFAR-100 with cosine annealing. 
Typical data augmentations including random shift and horizontal flip are applied.

\vspace{4pt}
\noindent\textbf{Parameters Setting.} For baselines, we use the same parameter settings in \citep{NCE, ALFs_ICML, ANL}, which match their best parameters. The detailed parameters for JAL and baselines can be found in Table \ref{tab: parameter}. For LT-APL \cite{student_loss}, we take results directly from the original paper.
For our method, we follow a principled strategy for parameter tuning: the range of $a$ can be initially estimated through theoretical guidance, and then selected from $[5, 10, 20, 30]$ based on experimental results. 


\begin{table*}[!h]
\centering
\centering
\caption{Parameter settings for different methods.}
\label{tab: parameter}
\begin{tabular}{c|cccc}
\toprule
\textbf{Parameter}    
& \textbf{CIFAR-10}         & \textbf{CIFAR-100}         & \textbf{WebVision}         & \textbf{Clothing1M}        \\
\midrule
CE                                                                             & -              & -               & -               & -               \\
FL ($\gamma$)                                                                           & (0.5)              & (0.5)               & -               & -               \\
GCE ($q$)                                                                         & (0.9)              & (0.7)               & (0.7)               & (0.6)               \\
SCE ($\alpha, \beta, A$)                            & (0.1, 1, -4)     & (6, 1, -4)        & (10, 1, -4)       & (10, 1, -4)       \\
NCE                                                                            & -              & -               & -               & -               \\
NCE+RCE ($\alpha, \beta, A$)                        & (1, 1, -4)       & (10, 0.1, -4)     & (50, 0.1, -4)     & (10, 1, -4)       \\
NCE+AUL ($\alpha, \beta, a, p$)                     & (1, 3, 6.3, 1.5) & (10, 0.015, 6, 3) & -               & -               \\
NCE+AGCE ($\alpha, \beta, a, q$)                    & (10, 4, 6, 1.5)  & (10, 0.1, 1.8, 3) & (50, 0.1, 2.5, 3) & (50, 0.1, 2.5, 3) \\
ANL-CE ($\alpha, \beta$)                          & (5, 5)           & (10, 1)           & (20, 1)           & (5, 0.1)          \\
ANL-FL ($\alpha, \beta, \gamma$) & (5, 5, 0.5)      & (10, 1, 0.5)      & (20, 1, 0.5)      & (5, 0.1, 0.5)     \\
JAL-CE ($\alpha, \beta, a$)                         & (1, 1, 30)       & (5, 1, 20)        & (50, 1, 30)    & (5, 0.1, 5)       \\
JAL-FL ($\alpha, \beta, a, \gamma$)                 & (1, 1, 30, 0.5)  & (5, 1, 20, 0.5)   & (50, 1, 30, 0.5)   & (5, 0.1, 5, 0.5)               \\

\bottomrule
\end{tabular}
\end{table*}

\subsection{Evaluation on Real-World Datasets}
\noindent\textbf{Experiment Setting for WebVision / ILSVRC12.}\quad For WebVision,  we use the mini setting \cite{mentornet}, which includes the first 50 classes of the google image subset. We train a ResNet-50 using SGD for 250 epochs with initial learning rate 0.4, nesterov momentum 0.9 and weight decay $3 \times 10^{-5}$ and batch size 256. The learning rate is multiplied by 0.97 after each epoch of training. All the images are resized to $224 \times 224$. Typical data augmentations including random shift, color jittering, and  horizontal flip are applied. 
We train the model on Webvision and evaluate the trained model on the same 50 concepts on the corresponding WebVision and ILSVRC12 validation sets.

\vspace{4pt}
\noindent\textbf{Experiment Setting for Clothing1M.}\quad For Clothing1M, we use ResNet-50 pre-trained on ImageNet similar to \citep{clothing1m}. All the images are resized to $224 \times 224$. We use SGD with a momentum of 0.9, a weight decay of $1 \times 10^{-3}$, and batch size of 256. We train the network for 10 epochs with a learning rate of $5 \times10^{-3}$ and a decay of 0.1 at 5 epochs. Typical data augmentations including random shift and horizontal flip are applied.

\end{document}